\documentclass[11pt]{article}
\usepackage{acl}
\usepackage{times}
\usepackage{latexsym}
\usepackage[T1]{fontenc}
\usepackage[utf8]{inputenc}
\usepackage{microtype}
\usepackage{inconsolata}
\usepackage{graphicx}
\usepackage{hyperref}
\usepackage{multirow} 
\usepackage{multicol}
\usepackage{mdframed}
\usepackage{amsmath}

\title{Are LLMs reliable? An exploration of the reliability of large language models in clinical note generation}

\author{Kristine Ann M. Carandang, Jasper Meynard P. Araña, \\ \textbf{Ethan Robert A. Casin, Christopher P. Monterola,} \\ 
\textbf{Daniel Stanley  Y. Tan, Jesus Felix B. Valenzuela, Christian M. Alis}\\
        Analytics, Computing \& Complex Systems Laboratory \\
        Asian Institute of Management}

\begin{document}
\maketitle
\begin{abstract}

Due to the legal and ethical responsibilities of healthcare providers (HCPs) for accurate documentation and protection of patient data privacy, the natural variability in the responses of large language models (LLMs) presents challenges for incorporating clinical note generation (CNG) systems, driven by LLMs, into real-world clinical processes. The complexity is further amplified by the detailed nature of texts in CNG. To enhance the confidence of HCPs in tools powered by LLMs, this study evaluates the reliability of 12 open-weight and proprietary LLMs from Anthropic, Meta, Mistral, and OpenAI in CNG in terms of their ability to generate notes that are string equivalent (consistency rate), have the same meaning (semantic consistency) and are correct (semantic similarity), across several iterations using the same prompt. The results show that (1) LLMs from all model families are stable, such that their responses are semantically consistent despite being written in various ways, and (2) most of the LLMs generated notes close to the corresponding notes made by experts. Overall, Meta's Llama 70B was the most reliable, followed by Mistral's Small model. With these findings, we recommend the local deployment of these relatively smaller open-weight models for CNG to ensure compliance with data privacy regulations, as well as to improve the efficiency of HCPs in clinical documentation.

\end{abstract}

\section{Introduction}
\label{sec1}

The capability of LLMs to produce text similar to human writing has led to research on their potential role in aiding clinical documentation. This led to the development of clinical note generation (CNG) tools designed to address extended working hours and healthcare provider (HCP) fatigue \citep{balloch_use_2024, biswas_intelligent_2024, giorgi-etal-2023-wanglab, heilmeyer_2024, moramarco-etal-2022-human, tung_2024}, issues which have persisted despite the adoption of electronic health records \citep{wu_evaluating_2024, zhang_characteristics_2022, ghatnekar_digital_2021, maas_care2report_2020, momenipour_balancing_2019, quiroz_challenges_2019}. Considering the legal and ethical responsibility of HCPs to write accurate clinical documentation \citep{mccoy_understanding_2024}, the reliability of these tools is critical.

LLM reliability is typically assessed using \textit{inter-prompt stability} which checks the consistency of responses when subjected to a variety of prompts designed to elicit the same response \citep{azimi_evaluation_2025, cheng_relic_2024, dentella_systematic_2023, kozaily_accuracy_2024, li_improving_2024, luo_factual_2024, wang_prompt_2024}. An alternative is to evaluate \textit{intra-prompt stability} by checking the consistency of responses in several iterations using the same prompt \citep{atil_llm_2024, barrie_prompt_2024, dentella_systematic_2023, savage_large_2024, saxena_evaluating_2024, yim_err_2024, zhao_improving_2024}. However, assessing LLM reliability is more challenging for natural language generation tasks, especially long-form text generation such as in CNG. Evaluation typically requires reference texts so that comparisons can be made using automatic evaluation metrics, and involves human evaluation as it remains the gold standard \citep{giorgi-etal-2023-wanglab, moramarco-etal-2022-human}. 

Although there exist studies that evaluated LLM performance in CNG from transcripts of provider-patient conversations \citep{balloch_use_2024, chen_exploring_2024, giorgi-etal-2023-wanglab}, only \citet{kernberg_using_2024} evaluated LLM reliability in CNG in terms of intra-prompt stability. While \citet{kernberg_using_2024} evaluated only one proprietary LLM, their findings show the variability in LLM responses. This may raise concerns on reliability if integrated in the clinical setting \citep{kernberg_using_2024}, similar with incorporating other healthcare tools developed using LLMs or artificial intelligence in general \citep{tucci_factors_2021, wang_applications_2024}. 

Additionally, no study exploring the reliability of open-weight LLMs in CNG was found. Using open-weight LLMs over proprietary ones is a typical consideration for healthcare applications due to data privacy concerns related to protected and sensitive health information \citep{giorgi-etal-2023-wanglab, heilmeyer_2024, wang2024adaptingopensourcelargelanguage}.

In this study, we sought to determine whether LLMs are reliable in CNG by evaluating how consistent and correct their generated notes are when using the same prompt in multiple iterations. We focus our evaluation on the CNG task of producing a clinical note based on a transcript of a conversation between a healthcare provider and a patient using an LLM. Four (4) proprietary models and eight (8) open-weight models from Anthropic, Meta, Mistral, and OpenAI were evaluated. This is done with the intention of providing evidence to HCPs on the reliability of LLMs. By doing so, we aim to enhance the body of knowledge regarding the design of reliable tools, crucial for industries such as healthcare, which would benefit from incorporation into real clinical workflows.

More concretely, our findings contribute to our continuous efforts to validate and improve the LLM-powered CNG component of SINTA (\textit{Scalable Intelligent Note-taking and Teaching-learning Assistant}), a system that we have developed to alleviate the workload of HCPs. Supported by an innovation grant from a government-run tertiary training hospital, we are currently evaluating our system with the goal of integrating it into their clinical workflows. 

\section{Related Work}

With the ability of LLMs to generate texts, recent work explored the performance of various ChatGPT models (i.e., ChatGPT 3.5 Turbo, ChatGPT 4) in CNG from transcripts of provider-patient conversations to assess the potential of using an LLM in ambient clinical documentation \citep{balloch_use_2024} or to compare the performance of fine-tuned pretrained encoder-decoder or decoder-only language models with at least an LLM \citep{chen_exploring_2024, giorgi-etal-2023-wanglab}.  \citet{kernberg_using_2024} assessed not only the correctness of notes generated from ChatGPT 4, but also the reliability of its responses through three repeated runs for each input, although they did not alter model parameters to make the model more deterministic. In addition, they used standardized assessments rated by human experts to evaluate the quality  of responses, without using automatic evaluation metrics. Aside from ChatGPT, we evaluate various open-weight and proprietary LLMs in generating clinical notes from provider-patient dialogues. 

Some studies \citep{atil_llm_2024, savage_large_2024, yim_err_2024} that evaluated LLM reliability also set model parameters that influence the determinism of LLMs, $temperature$, $top\_p$ and $top\_k$, to a value of or close to 0 to make the model deterministic. We also set the model parameters to make them more deterministic.

Assessing LLM performance in CNG usually requires reference notes against which LLM outputs are matched via automated evaluation metrics that assess string overlap or semantic similarity. These evaluations are frequently supplemented by human judgment \citep{giorgi-etal-2023-wanglab, moramarco-etal-2022-human}. \citet{giorgi-etal-2023-wanglab, moramarco-etal-2022-human} found BERTScore \citep{zhang2020bertscoreevaluatingtextgeneration}, an automatic evaluation metric that checks the similarity of two texts in the embedding space, to be the most appropriate embedding-based metric for the task of CNG. We use BERTScore to measure semantic consistency across responses per prompt and semantic similarity of the responses with the notes generated by experts.

In addition to semantic consistency and semantic similarity, we also measure consistency rate to reflect how much of its responses are string equivalent.  Consistency of responses was usually measured by considering string equivalence (\textit{total agreement rate for raw model response} \citep{atil_llm_2024}) or by semantic equivalence (\textit{consistency rate} \citep{zhao_improving_2024} or \textit{sample consistency} \citep{savage_large_2024}) in reliability evaluation studies. String equivalence was noted as a strict measure of reliability while evaluating whether responses contextually mean the same is specifically important in CNG due to stylistic differences of HCPs in documenting their sessions \citep{moramarco-etal-2022-human}.  

\section{Method}

\begin{figure*}[t]
        \centering
        \includegraphics[width=1\linewidth]{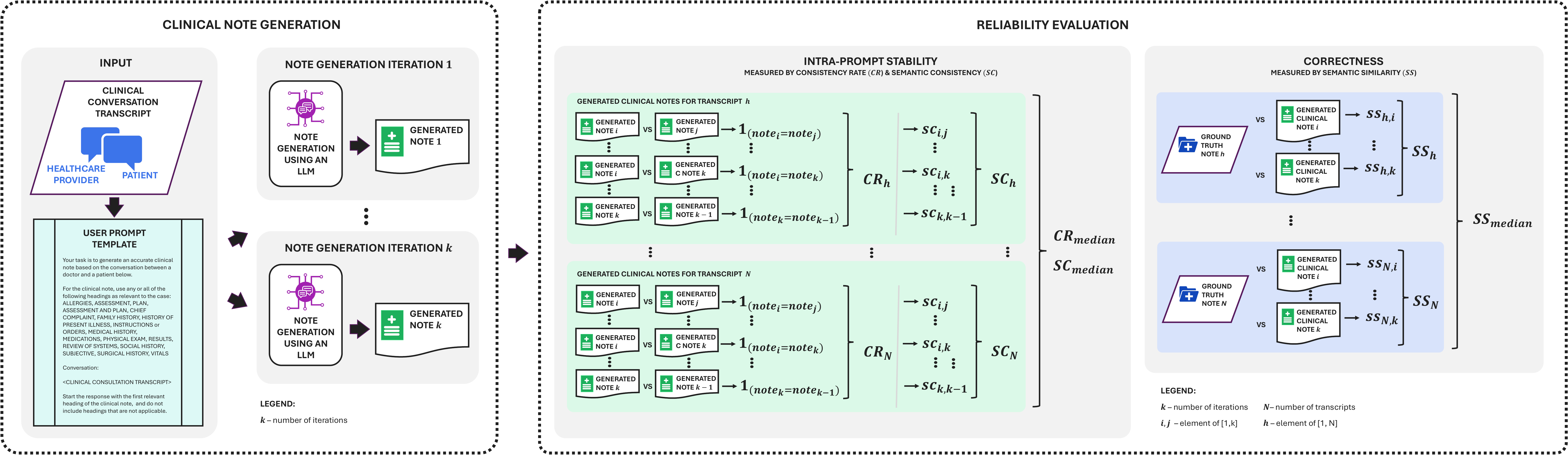}
        \caption{\textbf{Large Language Model (LLM) Reliability Evaluation Framework for the Task of Clinical Note Generation (CNG).} This has two phases, CNG and reliability evaluation, which are executed for each \textit{transcript} that has a corresponding clinical note made by an expert (\textit{ground truth note}). (1) \textit{CNG} starts with said transcript being incorporated into a \textit{user prompt template}, which then serves as an \textit{input} to an LLM. The LLM response is the \textit{generated note}. For each transcript, CNG is executed for $k$ iterations, resulting in $k$ generated notes. (2) \textit{Reliability Evaluation} is then done to assess LLM reliability according to its \textit{intra-prompt stability} and \textit{correctness}. \textit{Intra-prompt stability} is measured by consistency rate ($CR$) and median semantic consistency ($SC_{median}$), whereas \textit{correctness} is evaluated by its median semantic similarity ($SS_{median}$). Once done for all transcripts, model performance is calculated by taking the median of these scores.}
        \label{fig:framework}
\end{figure*}

We evaluate the reliability of LLMs according to their \textit{intra-prompt stability} and their \textit{correctness} following the process illustrated in Figure \ref{fig:framework}. Each transcript was incorporated into a user prompt template which instructs the LLM to generate a clinical note, with specified headings, from the transcript, for $k$ iterations. Evaluation was then done by using automatic evaluation metrics to determine consistency rate ($CR$) and semantic consistency ($SC$) as measures of intra-prompt stability, and correctness through its semantic similarity ($SS$).

\subsection{Dataset}
\label{sec:data}

We use \href{https://github.com/wyim/aci-bench}{\textbf{aci-bench}} \cite{yim_aci-bench_2023}, a benchmark dataset for automatic visit note generation, licensed under the Creative Commons Attribution 4.0 International Licence (CC BY). We select the $aci$ subset comprising 112 transcripts of natural conversations in English between a patient and a doctor taken during a role-play of a session, as this reflected the real-world scenario the most. Each data point contains the consultation session ID, the corrected transcript of the dialogue (\textit{transcript}), and the corresponding clinical note (\textit{ground truth note}). Figure \ref{fig:sample_data} in Appendix \ref{app:sample} shows an example of said transcript and ground truth note.

\subsection{Clinical Note Generation}
\label{sec:clin-note-gen}

Clinical notes were generated based on said \textit{transcripts} using the same prompt in multiple iterations using several LLMs configured to maximize determinism.

\subsubsection{User Prompt}
\label{sec:input}

A user prompt template was used across all iterations to have a consistent format for the input. This template (Figure \ref{app:user_prompt}), contains (1) the task, (2) the list of note headings present in the dataset, (3) the \textit{transcript}, and (4) other specific instructions. When using Llama models, modifications to this user prompt had to be made to align with the required format (see Appendix \ref{app:formatted_prompt}).

\subsubsection{Models \& their Configurations}
\label{sec:llms}

Various versions of open-weight LLMs (i.e, models from Meta and Mistral) and proprietary LLMs (i.e., models from Anthropic and OpenAI) were explored (Table \ref{tab:model_params}). These were accessed using AWS Bedrock API requests through the AWS SDK for Python (Boto3), except for OpenAI's models, which required the use of its API from its \href{https://openai.com/api/}{platform}.

\begin{table}[t]
    \centering
    \resizebox{\columnwidth}{!}{
    \begin{tabular}{llcccc}
        \hline
        \multirow{2}{*}{\textbf{Developer}} & \multirow{2}{*}{\textbf{Model}} & \multicolumn{4}{c}{\textbf{Model Configurations}} \\
        \cline{3-6}
         & &  \textbf{max output tokens} & \textbf{temperature} & \textbf{top\_p} & \textbf{top\_k}\\
         \hline
        Anthropic & Claude 3.5 Haiku & 8192 & 0 & 0 & 1 \\ 
          
         & Claude 3.5 Sonnet v2 & 8192 & 0 & 0 & 1\\ 
         
         \hline
        Meta & Llama 3.1-8B-Instruct & 2048 & 0 & 0 & N/A \\
         & Llama 3.1-70B-Instruct & 2048 & 0 & 0 & N/A \\ 
         & Llama 3.1-450B-Instruct & 8192 & 0 & 0 & N/A \\ 
         & Llama 3.2-1B-Instruct & 8192 & 0 & 0 & N/A \\ 
         & Llama 3.2-3B-Instruct & 8192 & 0 & 0 & N/A \\ 
         \hline
        Mistral & Large-2407 123B & 8192 & 0 & 0 & N/A \\
         & Small-2402 22B & 8192 & 0 & 0 & 1 \\
         & Mixtral-8x7B-Instruct & 4096 & 0 & 0 & 1 \\
        \hline
        OpenAI & ChatGPT-4o & 8192 & 0 & 0 & N/A \\
         & ChatGPT-40-mini & 8192 & 0 & 0 & N/A \\
         \hline
    \end{tabular}
    }
    \caption{\textbf{Models used and their parameters}. At least two LLM versions per developer was selected for use in this study - their smallest and their largest models. Maximum output tokens was set to 8192, unless otherwise specified due to model limitation. Other parameters were set accordingly to maximize determinism.}
    \label{tab:model_params}
\end{table}

At the minimum, for each model family, the smallest and largest models that took in multilingual text as input were included. Smaller models generally cost less than larger ones. For open-weight LLMs, smaller models also require less compute and storage resources than larger ones when deployed locally. Local deployment is an important option for CNG as this involves processing sensitive personal information which must be kept confidential in accordance with data privacy laws \citep{giorgi-etal-2023-wanglab, heilmeyer_2024, wang2024adaptingopensourcelargelanguage}. However, larger models were still considered, as they were generally reported to perform better in a variety of tasks than smaller models. 

For Meta's Llama 3.1 models, its 70B model was also considered in this study, as its largest model (405B) may be impractical to deploy in low resource settings. Llama 3.2 1B and 3B models were also included as they can be run locally on edge devices, which could more conveniently facilitate compliance with data privacy protection. 

Additionally, for the Mistral family, also included is their Mixtral model as this showcases a sparse mixture of experts model, which is said to improve computational efficiency compared with its counterpart LLMs. Not included in this study are the edge models of Mistral - Ministral 3B and 8B - as these were not available in AWS Bedrock at the time of the study.

To maximize determinism of these models during CNG,  three parameters known to influence model determinism, $temperature$, $top\_p$ and $top\_k$, were configured when relevant as enumerated in Table \ref{tab:model_params}. We also set the maximum output tokens to the respective maximum capacity of each model.

\subsection{Reliability Evaluation}
\label{sec:eval}

Reliability was assessed in terms of \textit{intra-prompt stability} and \textit{correctness} on ten (10) iterations to show how consistent an LLM generates notes across multiple runs using the same prompt, and how well an LLM generates notes compared to those made by experts, respectively. 

\subsubsection{Intra-Prompt Stability}
\label{sec:stability}

Intra-prompt stability is measured using the following metrics:

\begin{itemize}
    \item \textbf{\textit{Consistency Rate (CR)}} is measured by calculating the percentage of the number of pairs across the total number of iterations where a pair of generated notes are identical (i.e., string equivalent) over all possible combinations of pairs regardless of whether the outputs are correct. This strict measure of intra-prompt stability was first calculated per transcript ($CR_h$) as follows: 
    \begin{equation}
        CR_h = \frac{\sum_{i,j \in \binom{k}{2}} 1_{i=j}}{\binom{k}{2}}* 100
    \end{equation}
    where $k$ is the number of iterations and $h \in [1,N]$. Model performance was then calculated taking the median consistency rate ($CR_{median}$) from all $CR$ scores.
    \item \textbf{\textit{Semantic Consistency ($SC$)}} denotes whether the generated notes contextually mean the same regardless of how they were written across all iterations per transcript. This was measured by calculating BERTScore which is an automatic text generation evaluation metric that calculates the cosine similarity between a pair of notes in the contextual embedding space \citep{zhang2020bertscoreevaluatingtextgeneration}, using the implementation in \href{https://huggingface.co/spaces/evaluate-metric/bertscore}{Hugging Face}. The pairs of notes refer to all combinations of the ten (10) generated notes per transcript. To determine model performance, the median semantic consistency $SC_{median}$ was then calculated by getting the median of all semantic consistency scores calculated per transcript. 
\end{itemize}

\subsubsection{Correctness}
\label{sec:correctness}

Correctness is measured by \textbf{\textit{semantic similarity}}, which is similar to semantic consistency but the pair of notes compared here were the (1) generated note and (2) ground truth note made by an expert. The model performance ($SS_{median}$) was then calculated by taking the median of the semantic similarity scores calculated per transcript.

\section{Results and Discussion}
\label{sec3}

It took about 36 hours to generate the notes. Generally, we note that having perfect semantic consistency does not require having perfect consistency rate, and having perfect consistency rate and semantic consistency do not correspond to perfect semantic similarity as shown in Table \ref{tab:summary}.

\begin{table}[t]
    \centering
    \resizebox{\columnwidth}{!}{
    \begin{tabular}{llrrr}
        \hline
        \multirow{3}{*}{\textbf{Developer}} & \multirow{3}{*}{\textbf{Model}} & \multicolumn{3}{c}{\textbf{LLM Reliability $\uparrow$}}  \\
        \cline{3-5}
         & & \multicolumn{2}{c}{\textbf{Intra-prompt Stability}} & \textbf{Correctness} \\
         & &  \textbf{$CR_{median} \pm IQR $} & \textbf{$SC_{median} \pm IQR$ } & \textbf{$SS_{median} \pm IQR$ } \\
         \hline
        Anthropic 
        & Claude Haiku 3.5 & $\textbf{\underline{100.00}} \pm 00.00$ & $\textbf{\underline{100.00}} \pm 00.00$ & $85.61 \pm 0.97$  \\
        & Claude Sonnet 3.5 v2& $0.00 \pm 00.00$ & $96.86 \pm 1.44$ & $\underline{86.52} \pm 1.21$ \\         
         
         \hline
        Meta & Llama 3.1-8B & $35.56 \pm 41.11$ & $98.39 \pm 6.06$ & $83.71 \pm 3.17$  \\
         & Llama 3.1-70B & $80.00 \pm 37.78$ & $\underline{\textbf{100.00}} \pm 0.00$ & $85.90 \pm 1.29$  \\ 
         & Llama 3.1-450B & $22.22 \pm 20.00$ & $96.34 \pm 6.30$ & $\underline{86.72 \pm 1.95}$ \\ 
         & Llama 3.2-1B & $ \underline{\textbf{100.00}} \pm 00.00$ & $\underline{\textbf{100.00}} \pm 0.00$ & $80.49 \pm 2.53$  \\ 
         & Llama 3.2-3B & $\underline{\textbf{100.00}} \pm 00.00$ & $\underline{\textbf{100.00}} \pm 0.00$ & $83.85 \pm 1.39$ \\ 
         \hline
        Mistral & Large-2407 123B & $8.89 \pm 20.00$ & $97.75 \pm 2.83$ & $84.36 \pm 1.49$  \\
         & Small-2402 22B & $\underline{62.22} \pm 33.33$ & $\underline{\textbf{100.00}} \pm 0.00$ & $\underline{85.72} \pm 1.71$  \\
         & Mixtral-8x7B & $4.44 \pm 11.11$ & $96.14 \pm 4.64$ & $85.55 \pm 1.55$ \\
        \hline
        OpenAI & ChatGPT-4o & $0.00 \pm 00.00$ & $\underline{97.52} \pm 1.42$ & $87.01 \pm 1.33$ \\
         & ChatGPT-4o-mini & $0.00 \pm 00.00$ & $97.40 \pm 1.91$ & $\underline{\textbf{87.26}} \pm 1.24$ \\
         \hline
    \end{tabular}
    }
    \caption{\textbf{Summary of Model Performances based on Intra-Prompt Stability and Correctness}. In boldface are the best scores across all models while underlined are the best scores per model family. Three LLMs (25\%) demonstrated perfect consistency rate, 41.67\% ($n$=5) had perfect semantic consistency, and no model had perfect semantic similarity.}
    \label{tab:summary}
\end{table}

\subsection{Intra-prompt Stability}
\label{sec:result-stability}

\begin{figure}[t]
        \centering
        \includegraphics[width=1\linewidth]{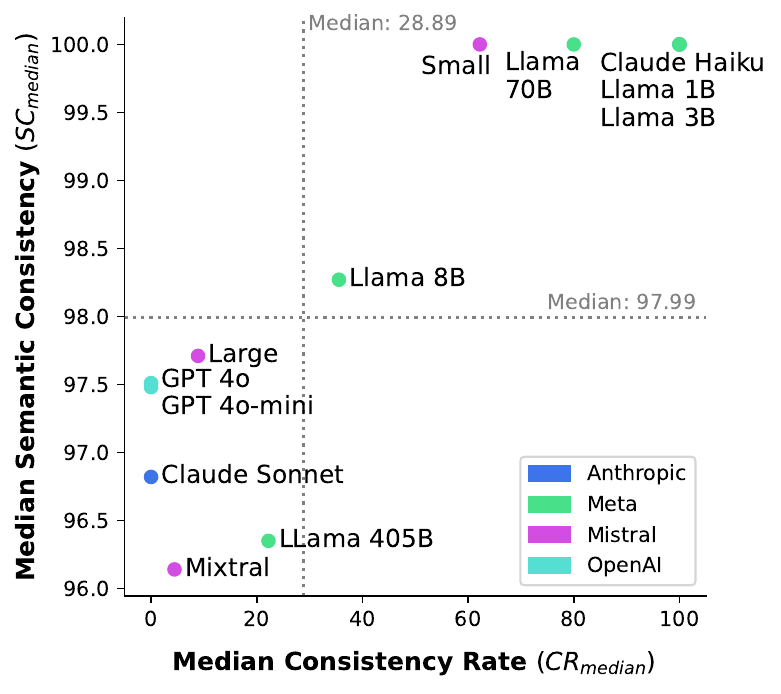}
        \caption{\textbf{Intra-prompt stability of LLMs in CNG.} Despite LLMs generating notes written in  varied ways, the meaning of these notes were relatively consistent across multiple iterations, implying that these LLMs performed well in terms of intra-prompt stability.
}
        \label{fig:stability}
\end{figure}

Figure \ref{fig:stability} shows the intra-prompt stability of LLMs in CNG. Meta's Llama 1B and 3B models, as well as Anthropic's Claude Haiku model, demonstrated perfect intra-prompt stability, which means that all outputs from all iterations were exactly the same and thus have the same meaning. Such performances seemed inconsistent with prior work on LLM intra-prompt stability evaluation for multiple choice question-answering tasks. Although the outputs of such tasks were linguistically controllable and are short-form texts, none of the LLMs studied by \citet{atil_llm_2024} had string equivalent responses in multiple executions using the same prompt in all questions, and no LLM had perfect semantic consistency in the study of \citet{zhao_improving_2024}.

Models from the Meta family generally performed better in terms of both consistency rate and semantic consistency than those from the other model families, whereas the models from the OpenAI family generally performed worse. Interestingly, for Anthropic, Meta and Mistral families, their smaller models performed remarkably better than their larger models.  Also worth noting are the performance of Meta's Llama 70B model and Mistral's Small model, which both had perfect semantic consistency despite having an imperfect, but notably high, consistency rate. 

In general, all models had a semantic consistency greater than 96\% regardless of the consistency rate, which varied greatly between models from 0\% to 100\%. This implies that despite the models generating clinical notes written in a variety of ways, the meaning of the content of these notes was relatively consistent across multiple iterations. Thus, all models performed well in terms of intra-prompt stability. This implies that intra-prompt stability may be measured using semantic consistency alone than with consistency rate. 

\subsection{Correctness}
\label{sec:result-correctness}

\begin{figure}[t]
        \centering
        \includegraphics[width=1\linewidth]{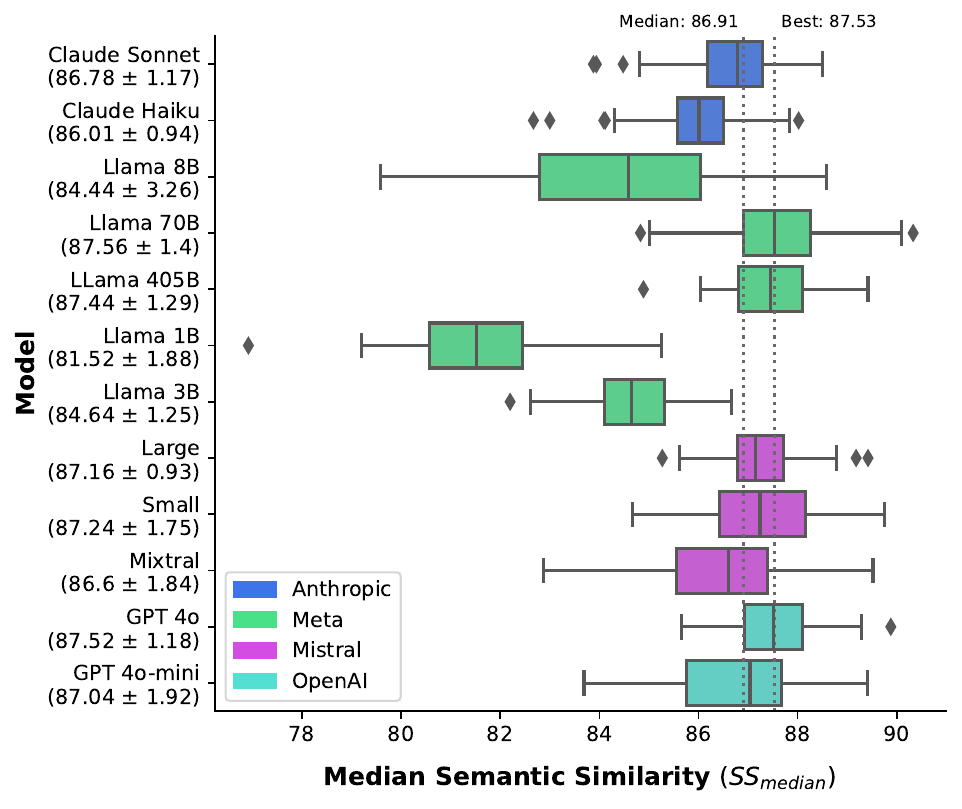}
        \caption{\textbf{Correctness of LLMs in CNG.} Except for Mistral, the larger models per model family performed better than their smaller models.}
        \label{fig:correctness}
\end{figure}

Figure \ref{fig:correctness} shows how close the generted notes were to the ground truth notes, indicating correctness. Generally, all LLMs had a median semantic similarity between 80 and 88. For Anthropic, Meta and OpenAI, their larger models performed better than their smaller models. For Mistral its Small model performed better than its Large model and its mixture-of-experts model.

A BERTScore of 80 is higher than the reported best performing LLM in the study of \citet{giorgi-etal-2023-wanglab}, which has a BERTScore of 60.8, as validated by senior resident physicians. Although they also used \textbf{aci-bench}, they incorporated in-context learning in their implementation with the $temperature$ parameter set to 0.2. 

\subsection{Overall LLM Reliability}

\begin{figure}[t]
        \centering
        \includegraphics[width=1\linewidth]{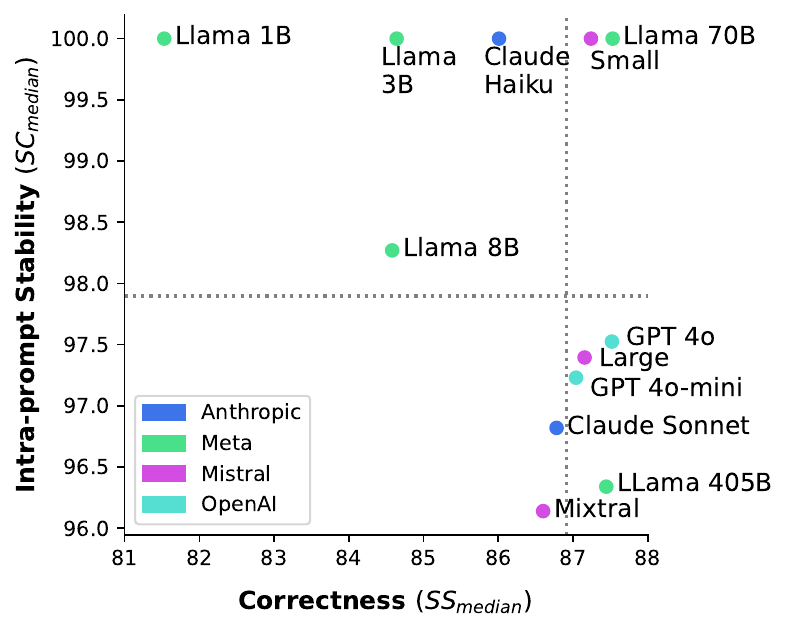}
        \caption{\textbf{Comparison of intra-prompt stability and correctness of LLMs in CNG.} Meta's Llama 70B model and Mistral's Small model appear to be among the most reliable models.}
        \label{fig:stability-correctness}
\end{figure}

Shown in Figure \ref{fig:stability-correctness} is the performance of the LLMs in CNG in terms of intra-prompt stability measured by semantic consistency and correctness measured by semantic similarity. Meta’s Llama 70B model performed the best considering both semantic consistency and semantic similarity, followed by Mistral’s Small model. These two open-weight models outperformed all proprietary models. For proprietary models, Anthropic's Claude Haiku had perfect semantic consistency but outperformed OpenAI's ChatGPT models in terms of semantic similarity. 

Using proprietary models through their respective platforms is more accessible to HCPs but may result in a breach of data privacy regulations as the prompts submitted may get added to their database and subsequently used for training. Furthermore, the parameters that maximize determinism cannot be configured in these platforms. With Meta's Llama 70B and Mistral Small outperforming the proprietary models, we can develop CNG tools that use these and make these more accessible to HCPs without data privacy issues.

\section{Conclusion}
\label{sec4}

The potential of LLMs for text generation has led to investigations into their ability to produce clinical notes, with the aim of improving the efficiency of documentation of HCPs. As part of our efforts to incorporate LLM-powered CNG tools into real clinical workflows, we have focused on building trust on these tools by assessing the reliability of LLMs in performing CNG.

Our observations indicate that LLMs do not consistently produce string-identical responses when aiming for semantically alike outputs, which are also aligned with annotations crafted by human experts. On multiple runs using the same prompt, we found that Meta's Llama 3.1 70B model was the most reliable, followed by Mistral's small model. Anthropic’s Claude Haiku model outperformed OpenAI’s ChatGPT 4o and 4o-mini models in terms of semantic consistency while the opposite was true for semantic similarity, but both proprietary models are subpar to Llama 3.1 70B and Mistral Small. With these findings, we recommend local deployment of these relatively smaller open-weight models for CNG to ensure compliance with data privacy regulations. We likewise consider using these models for SINTA as we validate its performance in the real world setting at the tertiary training hospital we are working with.

These findings provide support for the eventual integration of CNG tools powered by LLMs whilst protecting the health information of patients in compliance with data privacy regulations \citep{giorgi-etal-2023-wanglab, heilmeyer_2024, wang2024adaptingopensourcelargelanguage}. In this way, we can contribute to easing the burden of HCPs by providing them with tools that can help them comply with their documentation requirements more efficiently.

\section{Limitations}

As this study did not include prompt optimization, future work could involve comparing the same measures across various prompts to check for robustness and, at the same time, identify the prompt most suitable for the task. Metrics that utilize knowledge graphs and sentence parsers can also be used, along with an evaluation by human experts.

Also, our work only used one publicly available dataset that includes data gathered from simulations in English. We believe that it is necessary to conduct clinical validation and utility studies to capture and address contextual nuances before such tools can be fully adopted. 

\section{Ethical Considerations}

The data used includes transcripts of dialogues between HCPs and patients, taken from a publicly available dataset. Protected health information was not used.

Although we used proprietary models in our experiments such that the prompts we submitted may get added to their database and subsequently used for training, caution must be exercised when considering the use of these models in real clinical workflows to avoid any potential breach of data privacy regulations.

Since clinical note generation tools are being developed with the intent of being integrated in real clinical workflows, we recommend conducting clinical validation and clinical utility studies prior to integration to ensure that the tools meet health standards and comply with regulations. 

\bibliography{main}

\begin{thebibliography}{33}
\providecommand{\natexlab}[1]{#1}

\bibitem[{Atil et~al.(2024)Atil, Chittams, Fu, Ture, Xu, and Baldwin}]{atil_llm_2024}
Berk Atil, Alexa Chittams, Liseng Fu, Ferhan Ture, Lixinyu Xu, and Breck Baldwin. 2024.
\newblock \href {http://arxiv.org/abs/2408.04667} {{LLM} {Stability}: {A} detailed analysis with some surprises}.
\newblock \emph{arXiv preprint}.
\newblock ArXiv:2408.04667 [cs].

\bibitem[{Azimi et~al.(2025)Azimi, Qi, Wang, Rahmani, and Li}]{azimi_evaluation_2025}
Iman Azimi, Mohan Qi, Li~Wang, Amir~M. Rahmani, and Youlin Li. 2025.
\newblock \href {https://doi.org/10.1038/s41598-024-85003-w} {Evaluation of {LLMs} accuracy and consistency in the registered dietitian exam through prompt engineering and knowledge retrieval}.
\newblock \emph{Scientific Reports}, 15(1):1506.
\newblock Publisher: Nature Publishing Group.

\bibitem[{Balloch et~al.(2024)Balloch, Sridharan, Oldham, Wray, Gough, Robinson, Sebire, Khalil, Asgari, Tan, Taylor, and Pimenta}]{balloch_use_2024}
Jasmine Balloch, Shankar Sridharan, Geralyn Oldham, Jo~Wray, Paul Gough, Robert Robinson, Neil~J. Sebire, Saleh Khalil, Elham Asgari, Christopher Tan, Andrew Taylor, and Dominic Pimenta. 2024.
\newblock \href {https://doi.org/10.1016/j.fhj.2024.100157} {Use of an ambient artificial intelligence tool to improve quality of clinical documentation}.
\newblock \emph{Future Healthcare Journal}, 11(3):100157.

\bibitem[{Barrie et~al.(2024)Barrie, Palaiologou, and Törnberg}]{barrie_prompt_2024}
Christopher Barrie, Elli Palaiologou, and Petter Törnberg. 2024.
\newblock \href {https://doi.org/10.48550/arXiv.2407.02039} {Prompt {Stability} {Scoring} for {Text} {Annotation} with {Large} {Language} {Models}}.
\newblock \emph{arXiv preprint}.
\newblock ArXiv:2407.02039 [cs].

\bibitem[{Biswas and Talukdar(2024)}]{biswas_intelligent_2024}
Anjanava Biswas and Wrick Talukdar. 2024.
\newblock \href {http://arxiv.org/abs/2405.18346} {Intelligent {Clinical} {Documentation}: {Harnessing} {Generative} {AI} for {Patient}-{Centric} {Clinical} {Note} {Generation}}.
\newblock \emph{arXiv preprint}.
\newblock ArXiv:2405.18346.

\bibitem[{Chen and Hirschberg(2024)}]{chen_exploring_2024}
Yu-Wen Chen and Julia Hirschberg. 2024.
\newblock \href {https://doi.org/10.18653/v1/2024.clinicalnlp-1.1} {Exploring {Robustness} in {Doctor}-{Patient} {Conversation} {Summarization}: {An} {Analysis} of {Out}-of-{Domain} {SOAP} {Notes}}.
\newblock In \emph{Proceedings of the 6th {Clinical} {Natural} {Language} {Processing} {Workshop}}, pages 1--9, Mexico City, Mexico. Association for Computational Linguistics.

\bibitem[{Cheng et~al.(2024)Cheng, Zouhar, Arora, Sachan, Strobelt, and El-Assady}]{cheng_relic_2024}
Furui Cheng, Vilém Zouhar, Simran Arora, Mrinmaya Sachan, Hendrik Strobelt, and Mennatallah El-Assady. 2024.
\newblock \href {https://doi.org/10.1145/3613904.3641904} {{RELIC}: {Investigating} {Large} {Language} {Model} {Responses} using {Self}-{Consistency}}.
\newblock In \emph{Proceedings of the 2024 {CHI} {Conference} on {Human} {Factors} in {Computing} {Systems}}, {CHI} '24, pages 1--18, New York, NY, USA. Association for Computing Machinery.

\bibitem[{Dentella et~al.(2023)Dentella, Günther, and Leivada}]{dentella_systematic_2023}
Vittoria Dentella, Fritz Günther, and Evelina Leivada. 2023.
\newblock \href {https://doi.org/10.1073/pnas.2309583120} {Systematic testing of three {Language} {Models} reveals low language accuracy, absence of response stability, and a yes-response bias}.
\newblock \emph{Proceedings of the National Academy of Sciences}, 120(51):e2309583120.
\newblock Publisher: Proceedings of the National Academy of Sciences.

\bibitem[{Ghatnekar et~al.(2021)Ghatnekar, Faletsky, and Nambudiri}]{ghatnekar_digital_2021}
Shilpa Ghatnekar, Adam Faletsky, and Vinod~E. Nambudiri. 2021.
\newblock \href {https://doi.org/10.1007/s12553-021-00568-0} {Digital scribe utility and barriers to implementation in clinical practice: a scoping review}.
\newblock \emph{Health and Technology}, 11(4):803--809.

\bibitem[{Giorgi et~al.(2023)Giorgi, Toma, Xie, Chen, An, Zheng, and Wang}]{giorgi-etal-2023-wanglab}
John Giorgi, Augustin Toma, Ronald Xie, Sondra Chen, Kevin An, Grace Zheng, and Bo~Wang. 2023.
\newblock \href {https://doi.org/10.18653/v1/2023.clinicalnlp-1.36} {{W}ang{L}ab at {MEDIQA}-chat 2023: Clinical note generation from doctor-patient conversations using large language models}.
\newblock In \emph{Proceedings of the 5th Clinical Natural Language Processing Workshop}, pages 323--334, Toronto, Canada. Association for Computational Linguistics.

\bibitem[{Heilmeyer et~al.(2024)Heilmeyer, B{\"o}hringer, Reinhard, Arens, Lyssenko, and Haverkamp}]{heilmeyer_2024}
Felix Heilmeyer, Daniel B{\"o}hringer, Thomas Reinhard, Sebastian Arens, Lisa Lyssenko, and Christian Haverkamp. 2024.
\newblock \href {https://doi.org/10.2196/59617} {Viability of open large language models for clinical documentation in german health care: Real-world model evaluation study}.
\newblock \emph{JMIR Med Inform}, 12:e59617.

\bibitem[{Kernberg et~al.(2024)Kernberg, Gold, and Mohan}]{kernberg_using_2024}
Annessa Kernberg, Jeffrey~A. Gold, and Vishnu Mohan. 2024.
\newblock \href {https://doi.org/10.2196/54419} {Using {ChatGPT}-4 to {Create} {Structured} {Medical} {Notes} {From} {Audio} {Recordings} of {Physician}-{Patient} {Encounters}: {Comparative} {Study}}.
\newblock \emph{Journal of Medical Internet Research}, 26(1):e54419.
\newblock Company: Journal of Medical Internet Research Distributor: Journal of Medical Internet Research Institution: Journal of Medical Internet Research Label: Journal of Medical Internet Research Publisher: JMIR Publications Inc., Toronto, Canada.

\bibitem[{Kozaily et~al.(2024)Kozaily, Geagea, Akdogan, Atkins, Elshazly, Guglin, Tedford, and Wehbe}]{kozaily_accuracy_2024}
Elie Kozaily, Mabelissa Geagea, Ecem~R. Akdogan, Jessica Atkins, Mohamed~B. Elshazly, Maya Guglin, Ryan~J. Tedford, and Ramsey~M. Wehbe. 2024.
\newblock \href {https://doi.org/10.1016/j.ijcard.2024.132115} {Accuracy and consistency of online large language model-based artificial intelligence chat platforms in answering patients' questions about heart failure}.
\newblock \emph{International Journal of Cardiology}, 408:132115.

\bibitem[{Li et~al.(2024)Li, Li, and Zhang}]{li_improving_2024}
Taiji Li, Zhi Li, and Yin Zhang. 2024.
\newblock \href {https://aclanthology.org/2024.lrec-main.771/} {Improving {Faithfulness} of {Large} {Language} {Models} in {Summarization} via {Sliding} {Generation} and {Self}-{Consistency}}.
\newblock In \emph{Proceedings of the 2024 {Joint} {International} {Conference} on {Computational} {Linguistics}, {Language} {Resources} and {Evaluation} ({LREC}-{COLING} 2024)}, pages 8804--8817, Torino, Italia. ELRA and ICCL.

\bibitem[{Luo et~al.(2024)Luo, Xie, and Ananiadou}]{luo_factual_2024}
Zheheng Luo, Qianqian Xie, and Sophia Ananiadou. 2024.
\newblock \href {https://doi.org/10.1016/j.eswa.2024.124456} {Factual consistency evaluation of summarization in the {Era} of large language models}.
\newblock \emph{Expert Systems with Applications}, 254:124456.

\bibitem[{Maas et~al.(2020)Maas, Geurtsen, Nouwt, Schouten, van~de Water, van Dulmen, Dalpiaz, van Deemter, and Brinkkemper}]{maas_care2report_2020}
Lientje Maas, Mathan Geurtsen, Florian Nouwt, Stefan Schouten, Robin van~de Water, Sandra van Dulmen, Fabiano Dalpiaz, Kees van Deemter, and Sjaak Brinkkemper. 2020.
\newblock \href {https://core.ac.uk/reader/286030500} {The {Care2Report} {System}: {Automated} {Medical} {Reporting} as an {Integrated} {Solution} to {Reduce} {Administrative} {Burden} in {Healthcare}}.
\newblock \emph{Proceedings of the 53rd Hawaii International Conference on System Sciences}.

\bibitem[{McCoy et~al.(2024)McCoy, Ci~Ng, Sauer, Yap~Legaspi, Jain, Gallifant, McClurkin, Hammond, Goode, Gichoya, and Celi}]{mccoy_understanding_2024}
Liam~G. McCoy, Faye~Yu Ci~Ng, Christopher~M. Sauer, Katelyn~Edelwina Yap~Legaspi, Bhav Jain, Jack Gallifant, Michael McClurkin, Alessandro Hammond, Deirdre Goode, Judy Gichoya, and Leo~Anthony Celi. 2024.
\newblock \href {https://doi.org/10.1186/s12909-024-06048-z} {Understanding and training for the impact of large language models and artificial intelligence in healthcare practice: a narrative review}.
\newblock \emph{BMC Medical Education}, 24(1):1096.

\bibitem[{Momenipour and Pennathur(2019)}]{momenipour_balancing_2019}
Amirmasoud Momenipour and Priyadarshini~R. Pennathur. 2019.
\newblock \href {https://doi.org/10.1016/j.ergon.2019.06.012} {Balancing documentation and direct patient care activities: {A} study of a mature electronic health record system}.
\newblock \emph{International Journal of Industrial Ergonomics}, 72:338--346.

\bibitem[{Moramarco et~al.(2022)Moramarco, Papadopoulos~Korfiatis, Perera, Juric, Flann, Reiter, Belz, and Savkov}]{moramarco-etal-2022-human}
Francesco Moramarco, Alex Papadopoulos~Korfiatis, Mark Perera, Damir Juric, Jack Flann, Ehud Reiter, Anya Belz, and Aleksandar Savkov. 2022.
\newblock \href {https://doi.org/10.18653/v1/2022.acl-long.394} {Human evaluation and correlation with automatic metrics in consultation note generation}.
\newblock In \emph{Proceedings of the 60th Annual Meeting of the Association for Computational Linguistics (Volume 1: Long Papers)}, pages 5739--5754, Dublin, Ireland. Association for Computational Linguistics.

\bibitem[{Quiroz et~al.(2019)Quiroz, Laranjo, Kocaballi, Berkovsky, Rezazadegan, and Coiera}]{quiroz_challenges_2019}
Juan~C. Quiroz, Liliana Laranjo, Ahmet~Baki Kocaballi, Shlomo Berkovsky, Dana Rezazadegan, and Enrico Coiera. 2019.
\newblock \href {https://doi.org/10.1038/s41746-019-0190-1} {Challenges of developing a digital scribe to reduce clinical documentation burden}.
\newblock \emph{npj Digital Medicine}, 2(1):1--6.
\newblock Number: 1 Publisher: Nature Publishing Group.

\bibitem[{Savage et~al.(2024)Savage, Wang, Gallo, Boukil, Patel, Safavi-Naini, Soroush, and Chen}]{savage_large_2024}
Thomas Savage, John Wang, Robert Gallo, Abdessalem Boukil, Vishwesh Patel, Seyed Amir~Ahmad Safavi-Naini, Ali Soroush, and Jonathan~H. Chen. 2024.
\newblock \href {https://doi.org/10.1093/jamia/ocae254} {Large language model uncertainty proxies: discrimination and calibration for medical diagnosis and treatment}.
\newblock \emph{Journal of the American Medical Informatics Association: JAMIA}, page ocae254.

\bibitem[{Saxena et~al.(2024)Saxena, Chopra, and Tripathi}]{saxena_evaluating_2024}
Yash Saxena, Sarthak Chopra, and Arunendra~Mani Tripathi. 2024.
\newblock \href {https://doi.org/10.48550/arXiv.2404.16478} {Evaluating {Consistency} and {Reasoning} {Capabilities} of {Large} {Language} {Models}}.
\newblock \emph{arXiv preprint}.
\newblock ArXiv:2404.16478 [cs].

\bibitem[{Tucci et~al.(2021)Tucci, Saary, and Doyle}]{tucci_factors_2021}
Victoria Tucci, Joan Saary, and Thomas~E. Doyle. 2021.
\newblock \href {https://jmai.amegroups.org/article/view/6664} {Factors influencing trust in medical artificial intelligence for healthcare professionals: a narrative review}.
\newblock \emph{Journal of Medical Artificial Intelligence}, 5(0).

\bibitem[{Tung et~al.(2024)Tung, Gill, Sng, Lim, Ke, Tan, Jin, Elangovan, Ong, Abdullah, Ting, and Chong}]{tung_2024}
Joshua Yi~Min Tung, Sunil~Ravinder Gill, Gerald Gui~Ren Sng, Daniel Yan~Zheng Lim, Yuhe Ke, Ting~Fang Tan, Liyuan Jin, Kabilan Elangovan, Jasmine Chiat~Ling Ong, Hairil~Rizal Abdullah, Daniel Shu~Wei Ting, and Tsung~Wen Chong. 2024.
\newblock \href {https://doi.org/10.2196/57721} {Comparison of the quality of discharge letters written by large language models and junior clinicians: Single-blinded study}.
\newblock \emph{J Med Internet Res}, 26:e57721.

\bibitem[{Wang et~al.(2024{\natexlab{a}})Wang, Gao, Liu, Xu, Hussein, Labban, Iheasirim, Korsapati, Outcalt, and Sun}]{wang2024adaptingopensourcelargelanguage}
Hanyin Wang, Chufan Gao, Bolun Liu, Qiping Xu, Guleid Hussein, Mohamad~El Labban, Kingsley Iheasirim, Hariprasad Korsapati, Chuck Outcalt, and Jimeng Sun. 2024{\natexlab{a}}.
\newblock \href {https://arxiv.org/abs/2405.00715} {Adapting open-source large language models for cost-effective, expert-level clinical note generation with on-policy reinforcement learning}.
\newblock \emph{Preprint}, arXiv:2405.00715.

\bibitem[{Wang et~al.(2024{\natexlab{b}})Wang, Wan, Ni, Song, Li, Clayton, Malin, and Yin}]{wang_applications_2024}
Leyao Wang, Zhiyu Wan, Congning Ni, Qingyuan Song, Yang Li, Ellen Clayton, Bradley Malin, and Zhijun Yin. 2024{\natexlab{b}}.
\newblock \href {https://doi.org/10.2196/22769} {Applications and {Concerns} of {ChatGPT} and {Other} {Conversational} {Large} {Language} {Models} in {Health} {Care}: {Systematic} {Review}}.
\newblock \emph{Journal of Medical Internet Research}, 26(1):e22769.
\newblock Company: Journal of Medical Internet Research Distributor: Journal of Medical Internet Research Institution: Journal of Medical Internet Research Label: Journal of Medical Internet Research Publisher: JMIR Publications Inc., Toronto, Canada.

\bibitem[{Wang et~al.(2024{\natexlab{c}})Wang, Chen, Deng, Wen, You, Liu, Li, and Li}]{wang_prompt_2024}
Li~Wang, Xi~Chen, XiangWen Deng, Hao Wen, MingKe You, WeiZhi Liu, Qi~Li, and Jian Li. 2024{\natexlab{c}}.
\newblock \href {https://doi.org/10.1038/s41746-024-01029-4} {Prompt engineering in consistency and reliability with the evidence-based guideline for {LLMs}}.
\newblock \emph{npj Digital Medicine}, 7(1):1--9.
\newblock Publisher: Nature Publishing Group.

\bibitem[{Wu et~al.(2024)Wu, Wu, Wang, Lin, Liu, and Liu}]{wu_evaluating_2024}
Yuxuan Wu, Mingyue Wu, Changyu Wang, Jie Lin, Jialin Liu, and Siru Liu. 2024.
\newblock \href {https://doi.org/10.2196/54811} {Evaluating the {Prevalence} of {Burnout} {Among} {Health} {Care} {Professionals} {Related} to {Electronic} {Health} {Record} {Use}: {Systematic} {Review} and {Meta}-{Analysis}}.
\newblock \emph{JMIR Medical Informatics}, 12(1):e54811.
\newblock Company: JMIR Medical Informatics Distributor: JMIR Medical Informatics Institution: JMIR Medical Informatics Label: JMIR Medical Informatics Publisher: JMIR Publications Inc., Toronto, Canada.

\bibitem[{Yim et~al.(2023)Yim, Fu, Ben~Abacha, Snider, Lin, and Yetisgen}]{yim_aci-bench_2023}
Wen-wai Yim, Yujuan Fu, Asma Ben~Abacha, Neal Snider, Thomas Lin, and Meliha Yetisgen. 2023.
\newblock \href {https://doi.org/10.1038/s41597-023-02487-3} {Aci-bench: a {Novel} {Ambient} {Clinical} {Intelligence} {Dataset} for {Benchmarking} {Automatic} {Visit} {Note} {Generation}}.
\newblock \emph{Scientific Data}, 10(1):586.
\newblock Publisher: Nature Publishing Group.

\bibitem[{Yim et~al.(2024)Yim, Fu, Ben~Abacha, and Yetisgen}]{yim_err_2024}
Wen-wai Yim, Yujuan Fu, Asma Ben~Abacha, and Meliha Yetisgen. 2024.
\newblock \href {https://aclanthology.org/2024.lrec-main.1409} {To {Err} {Is} {Human}, {How} about {Medical} {Large} {Language} {Models}? {Comparing} {Pre}-trained {Language} {Models} for {Medical} {Assessment} {Errors} and {Reliability}}.
\newblock In \emph{Proceedings of the 2024 {Joint} {International} {Conference} on {Computational} {Linguistics}, {Language} {Resources} and {Evaluation} ({LREC}-{COLING} 2024)}, pages 16211--16223, Torino, Italia. ELRA and ICCL.

\bibitem[{Zhang et~al.(2020)Zhang, Kishore, Wu, Weinberger, and Artzi}]{zhang2020bertscoreevaluatingtextgeneration}
Tianyi Zhang, Varsha Kishore, Felix Wu, Kilian~Q. Weinberger, and Yoav Artzi. 2020.
\newblock \href {https://arxiv.org/abs/1904.09675} {Bertscore: Evaluating text generation with bert}.
\newblock \emph{Preprint}, arXiv:1904.09675.

\bibitem[{Zhang et~al.(2022)Zhang, Joy, Harris, and Park}]{zhang_characteristics_2022}
Zhan Zhang, Karen Joy, Richard Harris, and Sun~Young Park. 2022.
\newblock \href {https://doi.org/10.1145/3555111} {Characteristics and {Challenges} of {Clinical} {Documentation} in {Self}-{Organized} {Fast}-{Paced} {Medical} {Work}}.
\newblock \emph{Proceedings of the ACM on Human-Computer Interaction}, 6(CSCW2):386:1--386:21.

\bibitem[{Zhao et~al.(2024)Zhao, Yan, Sun, Xing, Wang, Meng, Cheng, Ren, and Yin}]{zhao_improving_2024}
Yukun Zhao, Lingyong Yan, Weiwei Sun, Guoliang Xing, Shuaiqiang Wang, Chong Meng, Zhicong Cheng, Zhaochun Ren, and Dawei Yin. 2024.
\newblock \href {https://aclanthology.org/2024.lrec-main.782} {Improving the {Robustness} of {Large} {Language} {Models} via {Consistency} {Alignment}}.
\newblock In \emph{Proceedings of the 2024 {Joint} {International} {Conference} on {Computational} {Linguistics}, {Language} {Resources} and {Evaluation} ({LREC}-{COLING} 2024)}, pages 8931--8941, Torino, Italia. ELRA and ICCL.

\end{thebibliography}

\newpage
\appendix

\section{Sample Data from aci-bench}
\label{app:sample}

Each data point of the \textbf{aci} subset of \textbf{ aci-bench} contains a corrected transcript of a natural conversation between a patient and a doctor (\textit{clinical conversation transcript}), together with its corresponding clinical note which serves as the \textit{ground truth note} for this study.

\begin{figure*}[ht]
    \centering
    \includegraphics[width=1\linewidth]{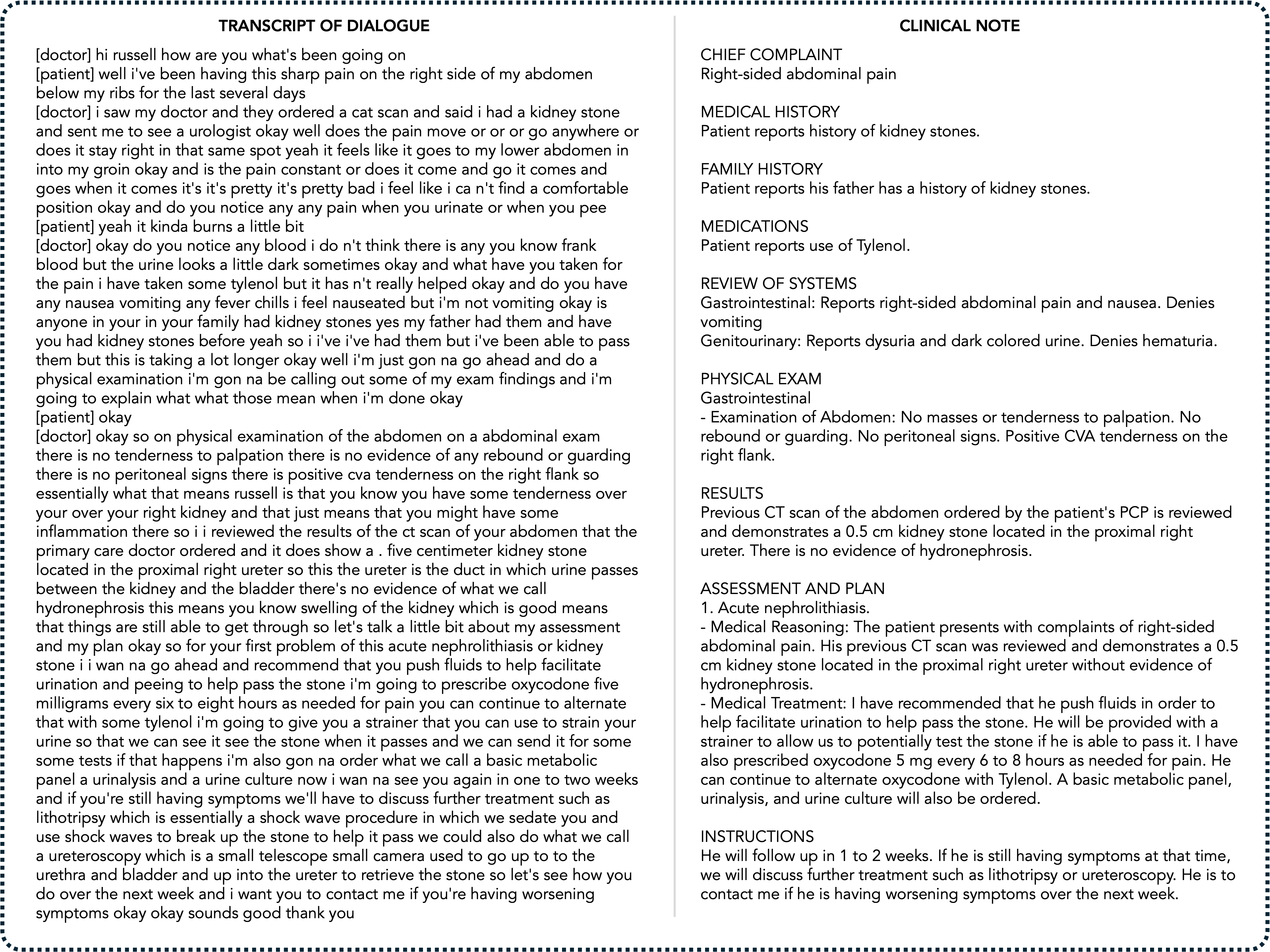}
    \caption{\textbf{Sample data from the \textit{aci-bench} dataset.} An example of the corrected transcript of a natural conversation between a patient and a doctor (\textit{clinical conversation transcript}), together with its corresponding clinical note which serves as the \textit{ground truth note} for this study.}
    \label{fig:sample_data}
\end{figure*}

\section{User Prompt Template }
\label{app:user_prompt}

Figure \ref{fig:user_prompt} shows the user prompt template used as input to the LLMs evaluated, except for Llama models. This template contains (1) the task, (2) the list of note headings present in the dataset, (3) the \textit{transcript}, and (4) other specific instructions.

\begin{figure*}[t]
    \centering
    \begin{mdframed}[nobreak=true]
    \small{
    \textbf{User Prompt Template}
    
    Your task is to generate an accurate clinical note based on the conversation between a doctor and a patient below. 
    
    For the clinical note, use any or all of the following headings as relevant to the case: ALLERGIES, ASSESSMENT, PLAN, ASSESSMENT AND PLAN, CHIEF COMPLAINT, FAMILY HISTORY, HISTORY OF PRESENT ILLNESS, INSTRUCTIONS or ORDERS, MEDICAL HISTORY, MEDICATIONS, PHYSICAL EXAM, RESULTS, REVIEW OF SYSTEMS, SOCIAL HISTORY, SUBJECTIVE, SURGICAL HISTORY, VITALS
    
    Conversation:
    
    $<clinical\ conversation\ transcript>$
    
    Start the response with the first relevant heading of the clinical note,  and do not include headings that are not applicable.  
    }
    \end{mdframed}
    \caption{\textbf{User Prompt Template.}  This was used to keep the format consistent across all models.}
    \label{fig:user_prompt}
\end{figure*}

\section{Formatted Prompt Template for Llama Models}
\label{app:formatted_prompt}

Llama models expect a certain format for the prompt, as shown in Figure \ref{fig:formatted_prompt}.

\begin{figure*}[t]
    \centering
    \begin{mdframed}[nobreak=true]
    \small{

    <|begin\_of\_text|> \\<|start\_header\_id|>user<|end\_header\_id|> \\
    {user\_prompt} \\
    <|eot\_id|>
    
    }
    \end{mdframed}
    \caption{\textbf{Formatted Prompt Template.} This was used to keep the format consistent across all Llama models. $user\_prompt$ here refers to the input which contains the transcript included in the User Prompt Template (Figure \ref{fig:user_prompt}).}
    \label{fig:formatted_prompt}
\end{figure*}

\end{document}